\pdfoutput=1

\documentclass[11pt]{article}

\usepackage[final]{acl}

\usepackage{times}
\usepackage{latexsym}

\usepackage[T1]{fontenc}

\usepackage[utf8]{inputenc}

\usepackage{microtype}

\usepackage{inconsolata}

\usepackage{graphicx}
\usepackage{amsmath}
\usepackage{booktabs}
\usepackage{array}

\usepackage{inconsolata}

\usepackage{graphicx}
\usepackage{booktabs} 
\usepackage{marvosym}  
\usepackage{graphicx}
\usepackage{caption}
\usepackage{stfloats} 
\usepackage{cuted}
\usepackage{multirow}
\usepackage{tabularx}

\usepackage{tocloft}
\usepackage{etoc}
\usepackage{hyperref}
    
\usepackage{titletoc}
\usepackage{pifont} 
\newcommand{\cmark}{\textcolor{green}{\ding{51}}}%
\newcommand{\xmark}{\textcolor{red}{\ding{55}}}%

\title{LongEval: A Comprehensive Analysis of Long-Text\\ Generation Through a Plan-based Paradigm}

\newcommand*\samethanks[1][\value{footnote}]{\footnotemark[#1]}
\author{
\textbf{Siwei Wu}\textsuperscript{1}\thanks{Equal Contribution.}\quad 
\textbf{Yizhi Li}\textsuperscript{1}\samethanks[1]\quad 
\textbf{Xingwei Qu}\textsuperscript{1}\quad
\textbf{Rishi Ravikumar}\textsuperscript{1}\quad
\textbf{Yucheng Li}\textsuperscript{2}\quad
\\
\textbf{Tyler Loakman}\textsuperscript{3}\quad
\textbf{Shanghaoran Quan}\textsuperscript{4}\quad
\textbf{Xiaoyong Wei}\textsuperscript{5}\quad
\textbf{Riza Batista-Navarro}\textsuperscript{1}\quad
\textbf{Chenghua Lin}\textsuperscript{1}\thanks{Corresponding Author.}\quad
\\
\textsuperscript{1}University of Manchester\quad
\textsuperscript{2}University of Surrey\quad 
\textsuperscript{3}University of Sheffield\quad\\
\textsuperscript{4}Peking University\quad
\textsuperscript{5}Hong Kong Polytechnic University\quad\\
\texttt{\{siwei.wu-2,chenghua.lin\}@manchester.ac.uk}
}

\begin{document}
\maketitle

\begin{abstract}
Large Language Models (LLMs) have achieved remarkable success in various natural language processing tasks, yet their ability to generate long-form content remains poorly understood and evaluated. 
Our analysis reveals that current LLMs struggle with length requirements and information density in long-text generation, with performance deteriorating as text length increases. 
To quantitively locate such a performance degradation and provide further insights on model development, we present \textbf{LongEval}, a benchmark that evaluates long-text generation through both \textit{direct} and \textit{plan-based} generation paradigms, inspired by cognitive and linguistic writing models. 
The comprehensive experiments in this work reveal interesting findings such as that while model size correlates with generation ability, the small-scale model (e.g., LongWriter), well-trained on long texts, has comparable performance.
All code and datasets are released in \url{https://github.com/Wusiwei0410/LongEval}.
\end{abstract}

\section{Introduction}

Large Language Models (LLMs) have revolutionized Natural Language Processing (NLP), achieving remarkable performance across a wide range of generation tasks including dialogue generation \cite{abdullin2024synthetic}, story creation \cite{zhao2023more}, open-ended text generation \cite{zhou2024balancing}, and complex reasoning task \cite{zhang2023chinese,wu2024comparative}. Although LLMs have been increasingly deployed in real-world applications, their ability to handle long-document generation remains underexplored despite their significance.

\begin{figure}[!tb]
\centering
\includegraphics[width=0.99\columnwidth]{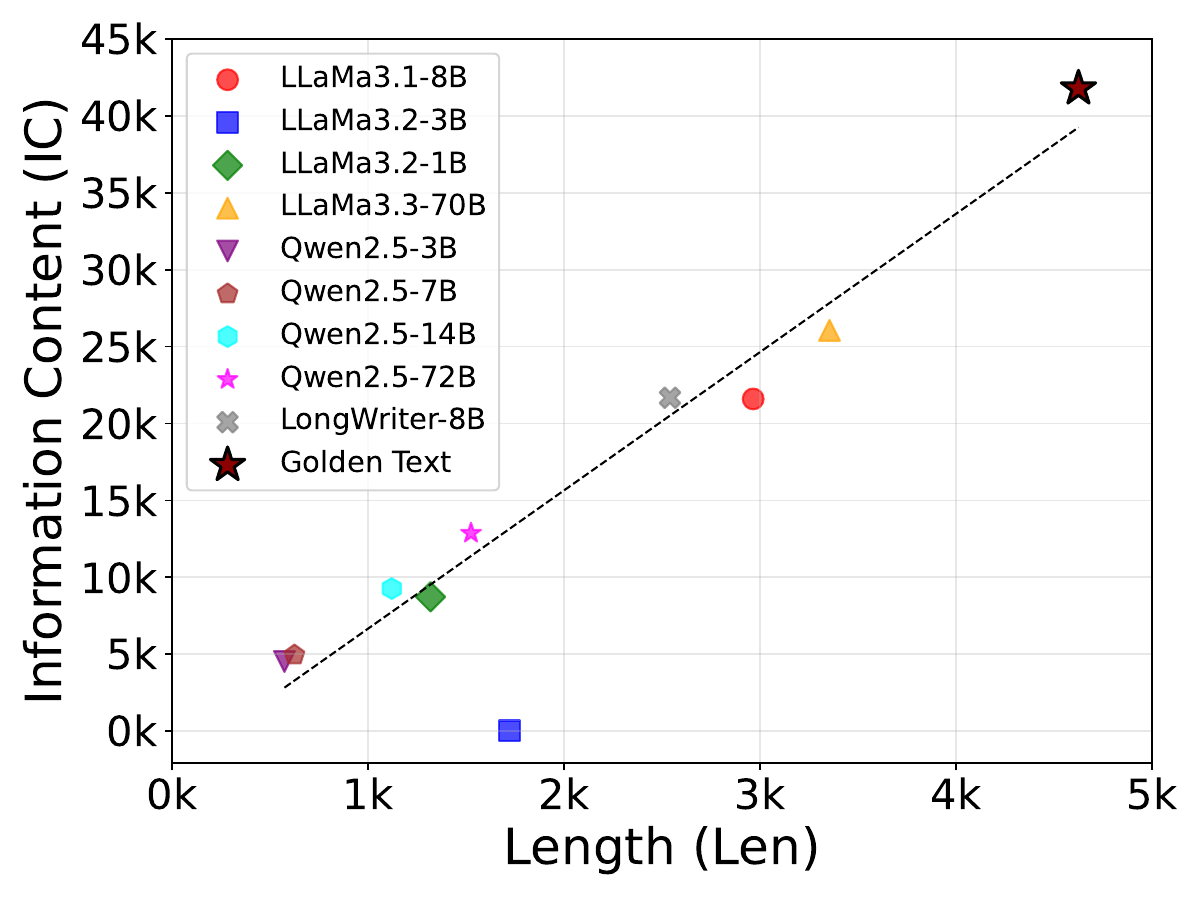}
\caption{The information content of LLMs-generated text and the golden human-authored text. We calculate information entropy using the frequency of each word in a document and determine the information content by multiplying the total word count by information entropy. 
}
\label{figure: information content}
\end{figure}

While there are recent studies seeking to improve the long-text generation ability \cite{bai2024longwriter,que2024hellobench} and long context understanding capability  \cite{xu2023retrieval,li2023loogle,li2024long,ding2024longrope,zhang2024bench}, the evaluation of long-text generation has been largely overlooked. Most existing benchmarks focus solely on long-context retrieval and understanding tasks \cite{bai2024longwriter,zhang2024longcite,pham2024suri,quan2024language,tang2024skyscript,an2024make}. A recent parallel work HelloBench \cite{que2024hellobench} proposes to evaluate the long-text generation by selecting samples from existing tasks (e.g., open-ended QA), where the tasks do not inherently require long generation capability.

\begin{figure*}[!tb]
    \centering
    \includegraphics[width=0.99\linewidth]{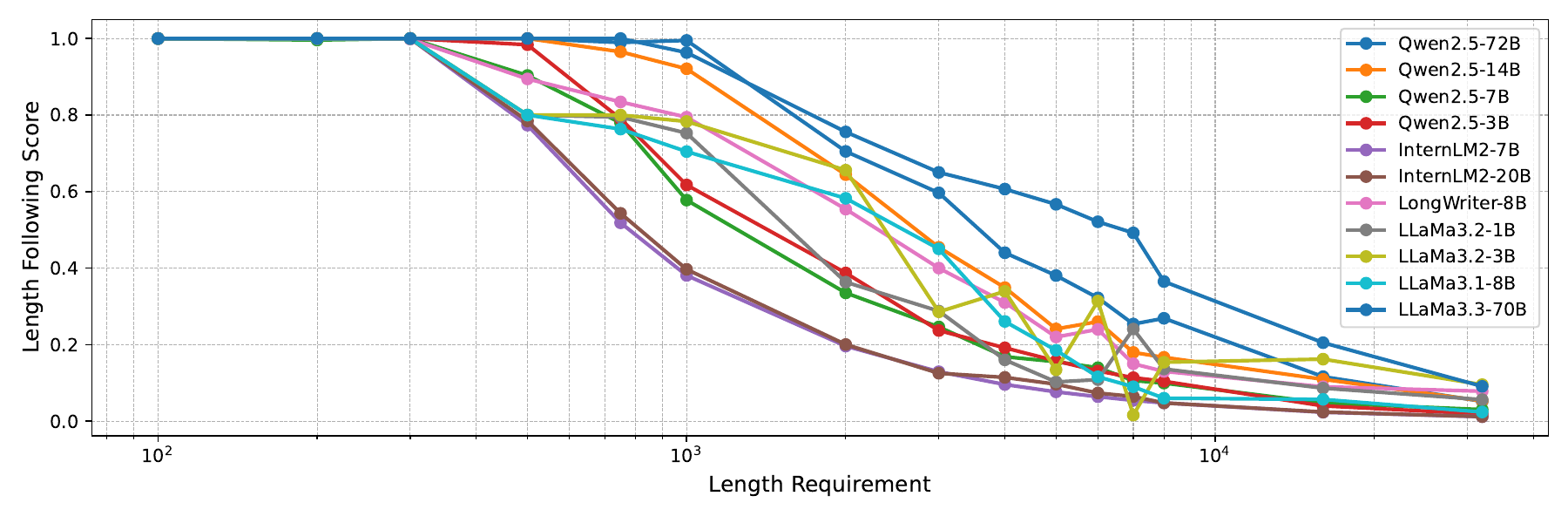}
    \caption{Th relation of the length requirement with the model-generated text length. Given the content plans, we require the LLMs to generate the text under various length requirements ranging from 100 to 32k. Specifically, we use the ratio of the generated text length to the requested length in the input as a score to evaluate the model's ability to follow length instructions.}
    \label{fig:motivation}
\end{figure*}

To comprehensively explore the long-generation capability of LLMs, we started with collecting a set of long and informative documents and using selected prevalent LLMs to directly reproduce the full documents from given summaries of those long documents. As shown in \autoref{figure: information content}, the information content in the documents is positively related to the length, which suggests the necessity of long text generation ability.
Furthermore, it could be observed that the prevalent LLMs (with parameters from 1B to 70B) still remain a large gap to the golden references regarding both information content and length dimensions.
We then tried to explore whether the LLMs could produce such long and informative documents by simply requiring to generate in specified lengths but failed.
LLMs tend to exhibit declining length-following abilities as the required length increases, with significant deterioration observed for texts exceeding 1k words, as revealed in \autoref{fig:motivation}.

Inspired by the cognitive writing theory, which posits that effective writing emerges from the process of ``\textit{cooking knowledge stored in long-term memory}'' through planning, translating, and reviewing~\citep{flower1980cognitivewriting}, we suspect that current generation paradigm of LLMs may be misaligned with human writing practices for long documents: \textit{LLMs often struggle to maintain consistency and provide deep insights in one-shot long-form writing, compared to plan-based writing}.
Specifically, the planning phase serves as a crucial foundation for developing coherent arguments and structured thoughts~\citep{scardamalia1987knowledge}, yet existing studies largely overlook this aspect of text generation.

To address these limitations, we introduce \textbf{LongEval}, a comprehensive benchmark designed to evaluate LLMs' long-text generation capabilities by supporting both direct and plan-based approaches. Our framework incorporates two key innovations: \emph{i)} a dual evaluation paradigm that assesses both zero-shot direct and plan-based structured generation that more closely align with human writing practices;
\emph{ii)} reliable automatic evaluation metrics that focus on content quality, structural coherence, and information density across various long text generation domains.

Since scientific texts and popular science articles often follow a prescribed writing structure, we select \textbf{three} long-text generation domains (i.e., \textbf{arXiv papers}, \textbf{blogs}, and \textbf{Wikipedia articles}) that necessitate that LLMs generate long-form texts (exceeding \textbf{2K words}) to build the benchmark for supporting a robust evaluation.
Different from similar work, HelloBench~\cite{que2024hellobench} (300 samples from general tasks) and LongWriter~\cite{bai2024longwriter} (120 synthetic samples for evaluation), we collect \textbf{166} high-quality human-authored samples that come from the long text generation domain.
We design a data production pipeline that leverages an advanced open-source LLM Qwen2.5-72B-Instruct\footnote{ \url{https://huggingface.co/Qwen/Qwen2.5-72B-Instruct}} to process documents from permissibly licensed sources across these different domains. In each document, sections are first summarized into comprehensive content as plans, with each major point elaborated in 4-5 sentences and verified by human annotators.

During the plan-based evaluation, the models are required to generate the full-text section-by-section using the summarized content plans as guidance, whilst required to maintain semantic consistency from previously generated sections. 
This approach systematically evaluates LLMs' long-text generation capabilities while aligning with the direct generation paradigm for sections.
Additionally, we design eight metrics to evaluate the generated long texts on different dimensions of quality. \emph{i)} To determine whether the LLM can follow instructions and whether the generated content is reasonable, we design the following domain-agnostic metrics at the \textbf{Document} level: Content-following (Cont-fol), Redundancy (Red), Length (Len), and Consistency (Con).
\emph{ii)} We design domain-specific metrics for the prescriptive domain of arXiv research papers that evaluate the following \textbf{sections}: Introduction (Intro), Related Work (RW), Method (ME), and Experimental Analysis (EA).

\section{Related Work}

\paragraph{Long Text Generation}
Recent research on long text generation has primarily focused on enhancing model performance~\cite{pham2024suri,zhang2024longcite,bai2024longwriter,quan2024language,tang2024skyscript,quan2024automatically}. A common approach involves constructing large-scale instruction-following datasets tailored for long-text generation and employing various optimization strategies to improve the capabilities of LLMs. 
Beyond direct model training, plan-based methods have gained traction for long-text generation. LongWriter~\cite{bai2024longwriter} demonstrates that synthetic datasets, generated using a structured planning approach with GPT-4o, can effectively enhance LLMs' ability to produce extended text. Similarly, \citet{wang2024autosurvey} propose a framework for generating survey papers section by section, while \citet{lu2024ai} employ a similar strategy to generate entire scientific articles. 
These studies suggest that structured generation methods can improve coherence and control over long-text outputs.

\paragraph{Long Context Understanding}
A key challenge in long-text generation is ensuring that LLMs effectively comprehend and utilize long contexts. Research in this area has focused on enhancing models' long-context understanding while extending their input length, leveraging their strong in-context learning capabilities~\cite{chen2023longlora,jiang2023longllmlingua,li-etal-2023-compressing,jin2024llm,zhang2024pqcache,ding2024longrope}. These efforts primarily target tasks such as reading comprehension, where models extract relevant information from lengthy inputs, as exemplified by benchmarks like LongICLBench~\cite{li2024long}, $\infty$BENCH~\cite{zhang2024bench}, and LonGLE~\cite{li2023loogle}. Despite these advancements, prior work has largely overlooked the challenge of generating coherent and contextually consistent long-form text beyond mere retrieval or summarization.

\paragraph{Long Text Evaluation}
Evaluating long-form text remains an open challenge. HelloBench~\cite{que2024hellobench} attempts to address this by selecting long-text samples of general tasks and evaluating LLMs through using direct generation method. Most existing evaluation frameworks rely on LLM-based scoring, but their robustness and reliability remain debated. As an alternative, \citet{zhang2024longreward} propose a reward model specifically designed for long-text evaluation. 

Additionally, several datasets have been developed to support long-text evaluation. Suri~\cite{pham2024surimulticonstraintinstructionfollowing} employs a plan-based approach and backtranslation~\cite{li2024selfalignmentinstructionbacktranslation, köksal2024longformeffectiveinstructiontuning} to generate instructional texts, though its focus is primarily on creative writing and blogs rather than academic content. In contrast, \citet{köksal2024longformeffectiveinstructiontuning} construct a long-text dataset based on Wikipedia and CommonCrawl, prioritizing direct text generation over structured planning. These studies highlight the need for high-quality datasets and evaluation metrics that account for both plan-based and direct-generation methods, particularly in domains requiring structured and coherent long-form outputs.

\begin{figure*}[!tb]
    \centering
    \includegraphics[width=0.99\textwidth]{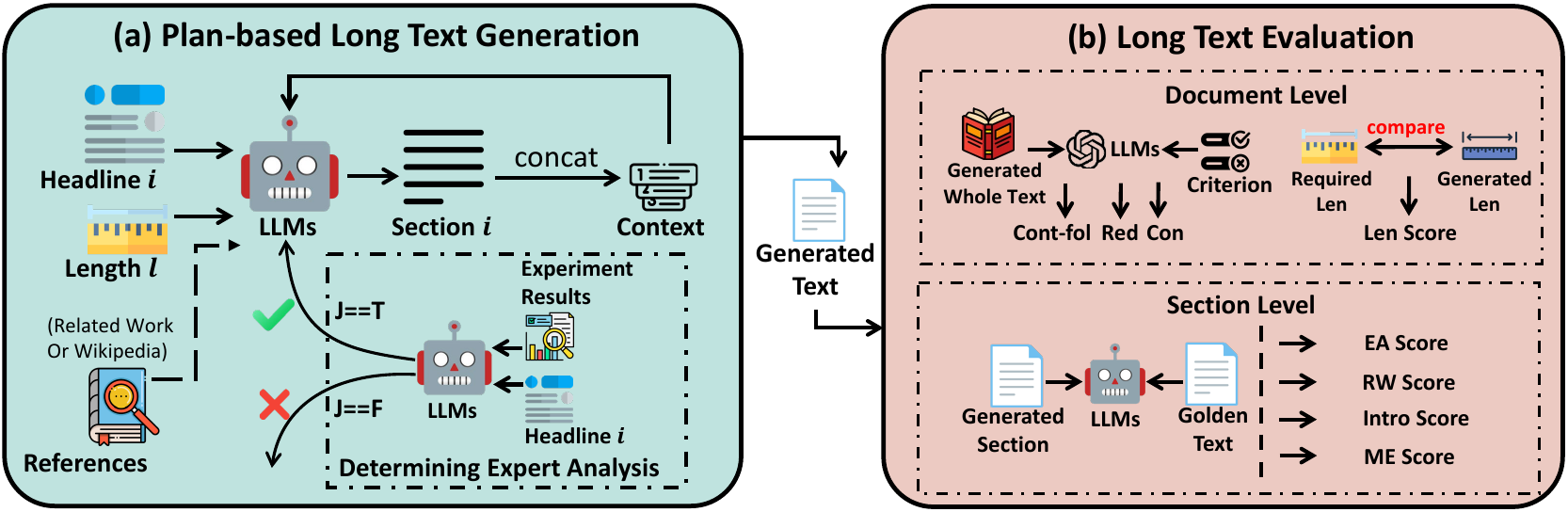}
    \caption{The Framework of our Long Text Generation method. Part (a) is the Plan-based method and part (b) is the Long Text Evaluation method.}
    \label{fig:framework}
\end{figure*}

\begin{table*}[h]
\centering
\footnotesize
\resizebox{1.95\columnwidth}{!}{
\begin{tabular}{l cccc}
\toprule
\multirow{2}{*}{\textbf{Benchmarks}} & \multicolumn{4}{c}{\textbf{Characteristics}} \\
\cmidrule(lr){2-5}
 & \textbf{Real Data} & \textbf{Plan Based} & \textbf{Domain Specific} & \textbf{Section \& Document Level} \\
\midrule
LongReward & \xmark & \xmark & \xmark & \xmark \\
LongWriter & \xmark & \cmark & \xmark & \xmark \\
HelloBench & \cmark & \xmark & \cmark & \xmark \\
LongEval (Ours) & \cmark & \cmark & \cmark & \cmark \\
\bottomrule
\end{tabular}
}
\caption{Comparison of different long-text generation benchmarks.}
\label{tab:method_comparison}
\end{table*}

\section{The LongEval Benchmark}
To fill the gap in the evaluation of long document generation, we propose \textbf{LongEval}, a benchmark built upon a unified framework for long-text generation, and introduce a comprehensive evaluation system.
Compared with similar studies, LongEval provides a robust evaluation system distinct across the dimension of data collection, generation paradigms, domain-specific and hierarchical metrics,
as shown in \autoref{tab:method_comparison}.
In this section, we first introduce a unified perspective of long text generation paradigms and then describe the accordingly designed evaluation systems.

\subsection{Long Text Generation Paradigms}
\label{sec: method}

The cognitive writing theory underscores the significance of planning in human writing~\cite{flower1980cognitivewriting}, and the plan-based paradigm has been effectively used to generate synthetic long-text data for training LLMs~\cite{bai2024longwriter}.
Therefore, generating ultra-long texts segment by segment is the mainstream paradigm~\cite{wang2024autosurvey,bai2024longwriter}. In this regard, this paper uses two methods (i.e., direct generation and plan-based generation) for long-text generation.

\paragraph{Direct Generation}
Although the direct generation method is applied to most NLP tasks, 
as shown in \autoref{fig:motivation}, most LLMs cannot directly generate text that exceeds 1k words.
In this work, we also evaluate the end-to-end long text generation capability of LLMs.
Specifically, we additionally perform direct generation by inputting the section content plan $p$, the article's length $l$, and other possible writing materials (e.g., experimental results $exp$, references $ref$) into LLMs.




\paragraph{Plan-Based Generation}
\label{subsec: Agent-Based Generation}

The plan-based methods are applied to generate long-length text due to its better performance than the direct method~\cite{bai2024longwriter,lu2024ai}.
Our experiments also analyze the length-following abilities of LLMs. To better understand the models' limitations, we conduct an in-depth investigation of LLM-generated content across different domains. 
\autoref{figure: information content} illustrates our quantitative analysis of the relationship between text length and information content, using human-written texts as a baseline. 
Therefore, as suggested by \autoref{fig:motivation}, we assume that current LLMs cannot meet the requirements of users who want to generate text with a large amount of information.
We design a unified plan-based generation method that uses the LLM to generate long text by section which ensures LLMs can generate text aligned with the length requirement.


As for each sample, we input the content plan $p$ of a section and the length requirement $l$ to make LLMs generate the whole article by section.
We additionally consider domain-specific writing requirements (e.g., for the arXiv paper domain, we use the experimental results as extra input to generate the results analysis section and for Wikipedia articles, we input the references to ensure the authenticity of the content). A detailed description of our plan-based generation method can be found in Appendix~\ref{sec: agent-based method}.

\subsection{Evaluation System and Prompts}
Previous works have primarily focused on studying the long-context understanding ability of LLMs~\cite{xu2023retrieval,li2024long,jin2024llm,zhang2024bench}. Most of these tasks resemble reading comprehension tasks and have standard answers (e.g., asking questions like `How old is Jack?' based on a long context).
Although HelloBench~\cite{que2024hellobench} has also evaluated the long-text generation ability of LLMs, their evaluation metrics do not take into account the characteristics of ultra-long text generation (such as the instruction-following ability in ultra-long text generation).
In this work, we evaluate the generation of long articles both at the \textbf{Document level} and the \textbf{Section level}.

\subsubsection{Domain-Agnostic Document-level Metrics}


\paragraph{Content-following (Cont-fol) Score.}
The input for generating long texts includes the writing outline (i.e., the content plan generated in \S\ref{subsec: sum headline}) of the entire article. Whether the model-generated text adheres to the requirements of the outline is a key factor in evaluating the quality of the generated text.
Therefore, as shown in \autoref{fig:prompt} in Appendix~\ref{sec:prompt}, we designed specialized prompts and input each section of the model-generated text along with the corresponding prompts to evaluate the model's ability to follow instructions for long-text generation.

\paragraph{Length-following (Len) Score}
For each section, we use the following method to calculate the length score:
\[
s =
\begin{cases} 
\frac{l_{\text{gen}}}{l_{\text{req}}}, & \text{if } l_{\text{gen}} < l_{\text{req}}, \\
1, & \text{otherwise}.
\end{cases}
\]
where $l_{gen}$ represents length of generated text, and $l_{req}$ represents length requirement in the prompt.
For section-level metrics, the final score is obtained by averaging the scores of all individual sections.

\paragraph{Redundancy (Red) Score.}
When generating long texts,  LLMs tend to treat each section as being independent, leading to potential redundancy across sections by repeating content. To address this, as shown in \autoref{fig:prompt},  we specifically designed a prompt to evaluate whether the content generated by the model contains redundant elements.

\paragraph{Consistency (Con) Score.}
For long-text writing, ensuring the connection between sections and paragraphs is crucial. Therefore, for model-generated text, as shown in \autoref{fig:prompt} in Appendix~\ref{sec:prompt}, we designed a prompt to evaluate its consistency.

\subsubsection{Domain-Specific Section-Level Metrics}
Due to some domains being more prescriptive in their format than others, we designed a range of evaluation criteria for the arXiv research paper and Wikipedia article domains that consider the expected structures of these more prescriptive formats.

\paragraph{Introduction (Intro) \& Related Work (RW) Scores.}
Since we provide a detailed writing outline and relevant references, we design a prompt to evaluate the Introduction and Related Work sections of arXiv papers, as shown in \autoref{fig:prompt} in Appendix~\ref{sec:prompt}. Using the original paper as the gold reference, we employed an LLM to assess the similarity between the generated text and the gold answer.
The blog writing format does not require the inclusion of references. While only papers contain specific related work sections, Wikipedia articles require extensive references throughout to ensure the authenticity of their content. Therefore, we treat the entire content of a Wikipedia article as a single related work section for evaluation.

\paragraph{Experiment Analysis (EA) Score.}
In the research paper domain, based on our observation, current LLMs struggle to determine which sections require the use of experimental results (e.g., they would use the results of the experiment in method). Furthermore, LLMs tend to merely reiterate the key points outlined without delving into the underlying reasons or connecting the causes behind different experimental results. Therefore, as shown in \autoref{fig:prompt} in Appendix~\ref{sec:prompt}, we design an evaluation prompt to compare the experimental analysis sections of the original article with those generated by the model.

\paragraph{Method (ME) Score.}
For method descriptions, the content generated by LLMs often consists of vague descriptions of methods without providing detailed design plans or formulaic explanations. To address this, as shown in \autoref{fig:prompt} in Appendix~\ref{sec:prompt}, we specifically designed a prompt to compare the method section of the original article with that generated by the model.

\section{Dataset Curation}
In previous studies~\citep{que2024hellobench}, one way to build the dataset for long-text generation evaluation is to filter long texts\footnote{The HelloBench study uses texts that are at least 1000 words long.} from existing tasks such as dialogue continuation.
Some of these tasks typically do not require long-text writing, making it difficult to fully assess the model's long-text generation capabilities in realistic scenarios.
Long-form content is prevalent across various domains, particularly in academic papers, blogs, and Wikipedia articles. Therefore, we construct a benchmark for long-text generation using data from these three domains to evaluate generation capabilities on naturally lengthy content.

\begin{table}[tb]
\begin{center} 
\footnotesize
\resizebox{0.99\columnwidth}{!}{
    \begin{tabular}{l|c|c|c|c}
\toprule
            & \textbf{GT\_len} & \textbf{Input\_len} & \textbf{ICR} & \textbf{Num} \\ \midrule
\textbf{arXiv}     & 4,754.28      & 1,038.46       & 21.84                    & 50              \\
\textbf{Wikipedia} & 3,323.54      & 844.09         & 25.40                    & 68              \\
\textbf{Blog}      & 2,623.10      & 766.19         & 29.21                    & 48              \\ \bottomrule
\end{tabular}
}
\end{center}
\caption{Data comparison across arXiv, Wikipedia, and blogs. IC presents Information Compression Ration.}
\label{fig:data_comparison}
\end{table}

\subsection{Data Collection Pipeline}
We design an automatic pipeline that collects documents from web pages without copyright restrictions and splits them into different sections according to predefined rules. We collect data from arxiv.org for papers, wikipedia.org for articles, and HuggingFace for blogs. These sources have permissible copyright licenses.
To ensure the quality of our benchmark, we hired one Postgrad student, who is familiar with the NLP, to manually check the processed data. Specifically, we delete the samples that do not follow a predefined format (e.g., a paper that does not have an abstract or a blog that misses an introduction). 

\subsection{Content Plan Generation}
\label{subsec: sum headline}

In order to support the plan-based long text-generation method introduced in \S\ref{subsec: Agent-Based Generation}, we use Qwen2.5-72B-Instruct to generate a content plan. Specifically, we pass each section of a document into the model and design a prompt to make the model summarize each section into 4-5 sentences. This forms the content plan for the section.

\subsubsection{Human Evaluation of Generated Content Plans}

\begin{table}[h]
\centering
\resizebox{0.95\columnwidth}{!}{
    \begin{tabular}{lcccc}
        \toprule
       & \textbf{arXiv} & \textbf{Wikipedia} & \textbf{Blog} & \textbf{Average} \\
        \midrule
        \textbf{Acc}&  86.2 & 88.6 & 91.4   & 88.7 \\
        \bottomrule
    \end{tabular}
}
\caption{The human evaluation results of LLM-summarized content plans.}
\label{fig:dataset_human_evaluation}
\end{table}

To assess whether the content plans preserve the key points of a document, we hire a postgraduate student to manually evaluate 10\% of the documents from each domain. Specifically, if the content plan for each section cannot capture sufficient relevant information, we regard it as an unqualified sample.
As shown in Table~\ref{fig:dataset_human_evaluation},
on \textbf{Wikipedia}, \textbf{Blog}, and \textbf{arXiv}, our manual evaluation accuracy is \textbf{88.6\%}, \textbf{91.4\%}, and \textbf{86.2\%}, respectively.
On average, \textbf{88.7\%} of the manually reviewed content plans contain adequate information, indicating that the content plans retain enough information for LLMs to faithfully (re)generate the content in the original document. 

\subsection{Dataset Characteristics}


As shown in Table~\ref{fig:data_comparison}, we analyze the average length of original samples (Ground Truth Length) and generate content plans across three domains. Among these domains, academic papers have the longest content plans, followed by Wikipedia articles and blogs. This pattern aligns with the inherent writing complexity of each domain: academic papers demand rigorous presentation, Wikipedia articles focus on popular science exposition, and blogs adopt a more informal style. This observation suggests a strong correlation between writing complexity and text length within each domain.

Our dataset maintains approximately 50 samples per domain, with the original text (ground truth) exceeding 2,000 words in each case. To evaluate the efficiency of our content plans, we introduced the Information Compression Ratio (ICR), defined as $\text{ICR} = L_{\text{GT}}/L_{\text{Input}}$, where $GT$ represents the ground truth text and $Input$ denotes the summarized content plan used as input for LLMs.
The ICR consistently ranges between 20\% and 30\% across all domains, indicating that,  to some extent, our content plans will not only retain the main content but also avoid disclosing too many details to the model.

\begin{table*}[hbt!]
\begin{center} 
\footnotesize
\resizebox{1.99\columnwidth}{!}{
    \begin{tabular}{l|l|c|ccccccccc}
\toprule
\textbf{Domain} & \textbf{Model} & \textbf{Overall} & \textbf{Intro} &\textbf{RW} & \textbf{EA} & \textbf{ME} & \textbf{Cont-fol} & \textbf{Len} & \textbf{Red} & \textbf{Con}
\\

\midrule
& GPT4o	 &81	&80&	79&	74	&79	&87	&93	&66	&84  \\

& Qwen2.5-3B-Instruct	&79&	80	&78	&75	&78	&84	&94	&67	&81 \\
&Qwen2.5-7B-Instruct&	80&	80	&79	&75	&78&	85&	93	&67	&83\\
&Qwen2.5-72B-Instruct	&\textbf{82}	&80&	78	&79	&79	&88&	94	&70&	84 \\
&Internlm2.5-7B-Chat&	71&	78	&78&	61	&65	&81	&75&	60&	75 \\
&Internlm2.5-20B-Chat&	73	&78&	78	&60&	57&	81	&75	&62&	76 \\
&Llama3.2-1B	&71	&78	&74&	60&	57	&71&	75	&72&	78 \\ 
&Llama3.2-3B	&76&	80&	78&66&	79	&73&	75&	72&	80\\
& Llama3.3-70B &79	&80	&80	&73	&86	&86	&97	&60	&82 \\
\multirow{-12}{*}{\textbf{arXiv}}&LongWriter-8B	&80&	80&	79	&77	&77&	86	&94&	68&	81 \\

\midrule
& GPT4o	&81	&78	&--&	--&	81	&83	&97	&68	&81	  \\
 &Qwen2.5-3B-Instruct	&80	&74	&--&	--&	77&	82&	74&	70&	77 \\
&Qwen2.5-7B-Instruct	& 81	&76&	--	&--	&82	&84	&76	&68	&80\\
&Qwen2.5-72B-Instruct&\textbf{83}	&75	&--	&--	&83	&84	&79	&71	&84\\
&Internlm2.5-7B-Chat&71	&76	&--	&--	&52	&68	&76	&66	&76	\\
&Internlm2.5-20B-Chat	&73	&77	&--	&--	&71	&62	&76	&67	&76 \\
&Llama3.2-1B	&70	&74	&--	&--	&55	&67	&75	&68	&74 \\
&Llama3.2-3B&79	&76	&--&--	&79	&75	&78	&76	&80 \\
& Llama3.3-70B &82	&78	&--	&--	&79	&86	&78	&66	&81\\
\multirow{-10}{*}{\textbf{Blog}}& LongWriter-8B	&\textbf{83}	&78	&--	&--	&82	&85	&79	&67	&84\\

\midrule
& GPT4o	&	81	&74	&80	&--	&85	&70	&95	&--	&82  \\
&Qwen2.5-3B-Instruct	&\textbf{82}	&75	&80	&--	&82	&71	&94	&--	&80\\
&Qwen2.5-7B-Instruct	&80	&75	&80	&--	&83	&67	&94&	--	&80 \\
&Qwen2.5-72B-Instruct	&81	&74&	80&	--&	84	&70	&94&--	&82\\
&Internlm2.5-7B-Chat&	71	&78	&77&	--	&69	&56	&90&	--&77\\
&Internlm2.5-20B-Chat	&73	&78	&77&	--	&71&	65	&90	&--	&76\\
&Llama3.2-1B	&71	&72	&71&	--&	68	&76	&67&	--	&72 \\
&Llama3.2-3B	&79&	80	&79	&--	&79	&76&	75&	--	&80\\
&Llama3.3-70B& \textbf{82}	&78	&80&	--	&84	&66	&99&	--	&81\\

\multirow{-10}{*}{\textbf{Wikipedia}}&	 LongWriter-8B	&\textbf{82}	&76	&81	&--	&85&	68	&98	&--	&82 \\

\bottomrule
\end{tabular}
}
\end{center}
\caption{
The plan-based results on our LongEval benchmark. We conduct experiments to evaluate current LLMs on three domains (i.e., arXiv papers, blogs, and Wikipedia articles). The `--' presents that the metric does not exist in this domain. The Overall is the average score of all indicators. For easier comparison, we retained only the integer part of all model scores.
}
\label{tab:Main Result}
\end{table*}

\section{Experiments and Result Analysis}
\label{sec: sec_experiments_results}

\subsection{Baseline}
We use a range of open-source LLMs, including \textbf{Llama3} (Llama3.2-1B, Llama3.2-3B, Llama3.3-70B)\cite{llama3modelcard}, \textbf{Qwen2.5} (3B, 7B, 72B)\cite{qwen2.5, qwen2}, and \textbf{InternLM2.5}, which excels in math reasoning~\cite{cai2024internlm2}. We also include \textbf{LongWriter}, a fine-tuned GLM model for long-form writing~\cite{bai2024longwriter}, and \textbf{GPT-4o}, a proprietary model with balanced performance across tasks.


\subsection{Overall Analysis}

\autoref{tab:Main Result} shows the experimental results of various models across the arXiv, Blog, and Wikipedia tasks.
The Qwen2.5 series models exhibit superior long-text generation capability, with Qwen2.5-72B-Instruct achieving the highest overall score of 82  in the arXiv domain and 83 in the Blog domain. It is followed by GPT-4o and LongWriter-8B. A consistent trend is observed where larger models within the same series outperform smaller ones, highlighting the benefits of scale in long-text generation.

Among the evaluation metrics, Cont-fol (Instruction Following) and Red (Redundancy) show the most significant performance differences. For instance, Qwen2.5-72B-Instruct scores 88 on Content-fol in the arXiv domain, while smaller models like InternLM2.5-7B-Chat achieve only 68. Similarly, in the Wikipedia domain, LongWriter-8B reaches 85, whereas InternLM2.5-7B-Chat lags at 69. These results suggest that instruction following and minimizing redundancy remain major challenges for long-text generation.
In contrast, RW, Intro, and Len have relatively smaller performance gaps. For example, across models in the arXiv domain, RW scores mostly cluster around 75-80, while, for most models, Len remains within 92-98. However, ME and EA  exhibit greater variation. Notably, in the arXiv domain, Qwen2.5-72B-Instruct scores 79 in ME, whereas InternLM2-5.7B-Chat only achieves 65. This suggests that while general writing ability remains relatively stable across models, tasks involving data analysis and experimental methodology pose greater challenges. When given structured writing guidance (e.g., content plans), models still struggle with high-level reasoning, requiring more advanced analytical capabilities to perform well.

\begin{table*}[hbt!]
\begin{center} 
\footnotesize
\resizebox{1.65\columnwidth}{!}{
    \begin{tabular}{c|c|lcccccccc}
\toprule
 \textbf{Random\_P} & \textbf{Overall} & \textbf{Con} &\textbf{RW} & \textbf{Intro} & \textbf{Len} & \textbf{EA} & \textbf{ME} & \textbf{Cont-fol} & \textbf{Red}
\\

\midrule

0.0 & 82   & 84  & 78 & 80  & 98 & 77 & 79 & 88 & 72 \\
0.1&   79  & 82 & 75 & 75  & 97 & 73 & 77 & 85 & 74 \\
0.2&   77  & 78 & 71 & 73  & 95 & 72 & 74 & 80 & 73 \\
0.3&    74  & 65 & 58 & 70  & 95 & 70 & 71 & 75 & 71 \\
0.5	&  72  & 57	&64 	&63 	&94 	&66& 	68 	&69 &	79 \\
0.7 &   69  & 54 & 63 & 56  & 95 & 64 & 64 & 62  & 75 \\
0.9 &   61   & 50  & 41 & 51  & 93 & 56 & 56 & 52 & 70   \\

\bottomrule
\end{tabular}
}
\end{center}
\caption{
The results of random replacement. 
}
\label{tab:metric Result}
\end{table*}

\begin{table}[tb]
\begin{center} 
\footnotesize
\resizebox{0.99\columnwidth}{!}{
    \begin{tabular}{l|l|c|c|c|c}
\toprule
     \textbf{Setting}    &   \textbf{Model}    & \textbf{Overall}	&	\textbf{Cont-fol} &		\textbf{Red}	&	 \textbf{Len}
 \\ \midrule
& \textbf{GPT4o} &61 &	82 &	82 &	21  \\
& \textbf{Qwen-3B} & 59 &	82 &	81 &	13 \\
& \textbf{Qwe-7B} & 60 &	81 &	85	& 15  \\
& \textbf{Qwen-72B} & 60 	& 84 &	40 &	58  \\
& \textbf{Llama-1B } & 52 &	71 	& 67 &	17 \\
& \textbf{Llama-3B} &58 &	78 &	69 	& 28            \\
& \textbf{Llama-70B} &63	&86 	&50 &	55          \\
& \textbf{IntLM2.5-7B} & 55 &	75 &	73 &	17
             \\
\multirow{-10}{*}{\textbf{Direct}}&	\textbf{IntLM2.5-20B} & 56 & 	75 &	75 &	18  \\  \midrule

& \textbf{GPT4o} &82  &	87  	&66 	&93    \\
&\textbf{Qwen-3B} & 81  &	84  &	67  &	94   \\
&\textbf{Qwen-7B} & 82   &	85  &	67  &	93     \\
&\textbf{Qwen-72B} &  86  &	88 &	72  &	98  \\
&\textbf{Llama-1B } & 73  &	71  &	72  &	75   \\
&\textbf{Llama-3B} &  79  &	79  &	70  &	89   \\
&\textbf{Llama-70B} & 81  &	86  &	60  &	97   \\
&\textbf{IntLM2.5-7B} & 71  &	78  & 	60  &	75    \\
 \multirow{-8}{*}{\textbf{Plan}}&	\textbf{IntLM2.5-20B} &  72  &	81 &	62 &	75  \\ \bottomrule

\end{tabular}
}
\end{center}
\caption{A comparison of direct and plan-based methods on domain-agnostic criteria. We use the arXiv domain subset only, owing to computational constraints.}
\label{tab:direct result}
\end{table}

\subsection{Long Text Generation Under Different Paradigm}\label{sec:compare_direct_and_plan}
As shown in \autoref{tab:direct result}, we compare the results of LLMs' long text generation ability under direct and plan-based settings. Notably, the overall score of the text generated by the plan-based method is much higher than that of Direct generation.
Additionally, we found that the text generated by the direct generation method is not only relatively short but also has a high level of redundancy.
This further proves the effectiveness of the plan-based generation method we designed and the plan-based method is more suitable for long text generation

\subsection{Effectiveness of LLM-As-A-Judge}

To validate the capability of LLM-as-a-judge of the LLMs on our metrics, we designed a random replacement test on the arXiv task where we randomly replaced $p\%$ sections in the model-generated content with sections sampled from other model-generated text and checked whether our model can identify the quality degradation and reflect it on the actual score. The test uses Qwen-2.5-72B's result with the $p$ from 0.1 to 0.9. As shown in \autoref{tab:metric Result}, for Instruction-following (Cont-fol), as the proportion of random replacements increases, the model's score drops sharply (from 88\% to 52\%).
For other metrics evaluating the quality of a specific section (RW, Intro, EA, ME), their scores also decrease overall as the proportion of random replacements increases.
This demonstrates that the scoring model can effectively identify changes in the content and quality, as well as reflect the content plan. As for the Length (Len) and Redundancy (Red) scores, they do not evaluate the content relevance between the generated text and the instruction but instead assess the quality of certain writing features within the text itself. As $p$ increases, Len and Red do not change significantly, indicating the robustness of this metric.

In addition, we also use GPT-4o as a judge model within our framework, as shown in \autoref{tab:GPT4o Result}. Although there are some differences in scores given by GPT-4o and Qwen2.5-72B on certain metrics, the distribution of scores between different models remains consistent. It demonstrates that Qwen2.5-72B also can effectively assess the long-text generation capabilities of LLMs under our framework.

\subsection{The Length Following Ability of LLMs}
To assess the ability of LLMs to generate texts of specified lengths, we directly instruct the models to produce texts of a specific length and compare the difference between the target length and the actual length (i.e., the Len metric). 
As shown in \autoref{fig:motivation}, our LLMs generate text with various length requirements ranging from 100 to 32,000 words. Most models achieve a Len Score of 1 when the required length (len\_req) is below 400. However, as len\_req increases, the Len Scores of all models decline sharply. When len\_req exceeds 4,000, most models score below 0.4, indicating that current LLMs struggle to generate long texts with precise length control. Notably, Qwen2.5 and Llama3 outperform other models, and larger models demonstrate stronger length-following capability.

\section{Conclusion}

The current long-text evaluation method overlooks long-text generation paradigms and lacks high-quality samples (e.g., the human-written text for the long-text generation task, such as paper writing).
In this work, we design a LongEval benchmark, collecting 156 long-text samples covering three domains that require the LLMs' long-text writing ability. 
We conduct experiments on mainstream LLMs and prove that the plan-based long-text generation method is more excellent than the direct generation method. Besides, although LLMs have a relatively better content-following ability, they still struggle with high-level reasoning writing (e.g., writing experiments analysis and designing methods).


\section*{Limitations}

Although the experiment result is significant, we only tested these models' performance in the arXiv domain under the direct setting to compare with the plan-based paradigm due to resource and time constraints.
In the future study, the benchmark should be considered to extend with the same data curation pipeline to achieve a more robust evaluation.

\section*{Ethics Statement}
The dataset used in our research is constructed using publicly available data sources, ensuring that
there are no privacy concerns or violations. We do not collect any personally identifiable information, and all data used in our research is obtained
following legal and ethical standards.
In the stage of data annotation, we employed three graduate students experienced in the Natural Language Processing field. We paid the graduate students approximately \$13 per hour, well above the local average wage, and engaged in constructive discussions if they had concerns about the process.

\bibliography{acl_latex}

\clearpage
\appendix
\section{Evaluation Prompts}
\begin{figure*}[!tb]
    \centering
    \includegraphics[width=0.99\textwidth]{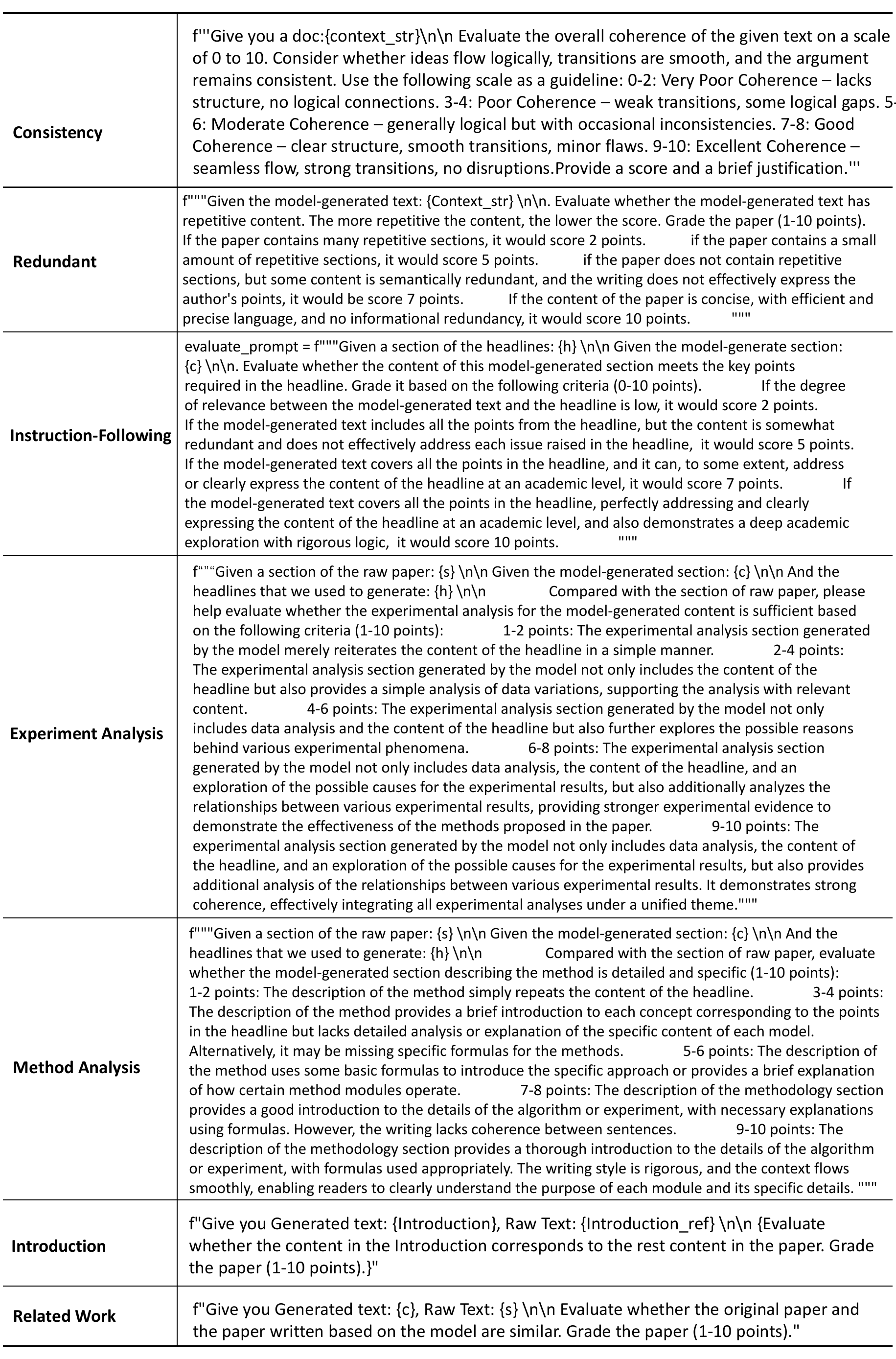}
    \caption{The table presents the prompts for the metrics that use LLMs to evaluate long text under different dimensions.}
    \label{fig:prompt}
\end{figure*}

We present the prompts that we designed for different long text generation dimensions in Tab~\ref{fig:prompt}.

\label{sec:prompt}

\section{Agent-based Generation Method}
\label{sec: agent-based method}

\paragraph{First section.}
We directly use the content plan $h$ and length $l$ to let the LLMs to generate the introduction of the article:

$$s=LLM(p,l,prompt),$$
where the $s$ is generated section.
Then we regard the $s$ as the context $c$.


\paragraph{Rest section.}
In the process of writing an article, it is often necessary to adjust the subsequent content based on the previous content.
Therefore, apart from the content plan $p$ and length requirement $l$, we also need to generate subsequent sections based on the previously generated context $c$ to ensure semantic consistency throughout the entire paper:

$$s = LLM(p,l,c,prompt),$$
then we concatenate $s$ and $c$ together as the context for generating subsequent sections.

\paragraph{Related work.}
As for the related work section of a paper, the LLM needs to use the references to write the background and development of the research direction. Besides, wikepedia docment also has to use numerous references to support the facility of the article.
Therefore, we input the extra reference $ref$ to generate the section:
$$s = LLM(h,l,c,ref,prompt)$$

\paragraph{Experiment analysis.}
As for the paper, there are many experiment analyses in different sections and there are no have obvious features in the subtitle of each section.
According to human writing behaviors, we input the content plan $p$ of the section and all the experiment results $exp$ of a paper into an LLM and let it judge whether they need to use the experiment results to write the content of the section:

$$judge = LLM(p,exp,prompt)$$

If the $judge$ is true, we will input the experiment results $res$ to have LLMs generate the current section, conversely, our generation strategy remains unchanged:
\[
s =
\begin{cases} 
LLM(p,l,c,exp,prompt), \text{if } judge {==} \text{T}, \\
LLM(p,l,c,prompt), \text{else }.
\end{cases}
\]

\paragraph{Final Result.} We contcat all the $s$ generated by our plan-absed method as the final generated articl $S$.

\section{The Evaluation Result by Using GPT4o }
\begin{table*}[hbt!]
\begin{center} 
\footnotesize
\resizebox{1.9\columnwidth}{!}{
    \begin{tabular}{l|l|l|ccccccccc}
\midrule[1pt]
\textbf{Domain} & \textbf{Model} & \textbf{Overall} & \textbf{Con} & \textbf{RW} & \textbf{Intro} & \textbf{Len} & \textbf{EA} & \textbf{ME} & \textbf{Cont-fol} & \textbf{Red}
\\

\midrule[1pt]

& Qwen-3B-Instruct	&75		&89		&82 	&63 		&58		&93	 &	93 		 	&51 	&85

 \\
&Qwen-7B-Instruct&		77 &	88	& 81 		&	73 		&64 	&	98 	 	&95 	&	40 		&85  

\\
&Qwen-72B-Instruct	&77		&87 	&78		&73 	&	68 	&	97 			&98 &	37		&88

 \\
&Internlm2.5-7B-Chat&		63 	&	86 &	81&		43 		&46 	&	76 	&	81 		&27		&76  

 \\
&Internlm2.5-20B-Chat&		68& 		86& 	81	&	53 		&52 	&	85 &		81 	&	40 	&	79  

 \\
 
& LLaMa3.3-70B &70	&	90	&85			&60		&61		&95		&62	&	39	&85

  \\
\multirow{-7}{*}{\textbf{arXiv}}&LongWriter-8B	&79	&	80& 	69	&	77	 &	77 		&86 	&		94& 	68	&	80

 \\

\midrule[1pt]
 &Qwen-3B-Instruct	& 75	&	84 	&	--		&63 	&	90 	&	45	&--		&95&80	

 \\
&Qwen-7B-Instruct	 &77	&88&	--		&	62 		&96 	&	44	&	--	&	98	&80

	 \\
&Qwen-72B-Instruct&  80&	84&	--	&		73 	&	97 	&	47 		&	--&	99	 &82

 \\
&Internlm2.5-7B-Chat& 63	&87	&	--		&	42	 &	70 		&31 &		--	&	84& 	74

	 \\
&Internlm2.5-20B-Chat	& 69	&	89	&--&	58& 		82	&	32	& 	--	&	 	84 &	 80

 \\

& LLaMa3.3-70B & 72	&	87 &	--&	60	&	87 &		26 		&	--	&	100	&84

 \\
\multirow{-7}{*}{\textbf{Blog}}& LongWriter-8B	& 77	& 87 &		--	&	69	&	96 	&	37 	&--	&	99 	&	84

\\

\midrule[1pt]

&Qwen3B-Instruct	& 79 &		84 	 &74	 &	--		 &--	 &	94 		 &	95 & 	49	  &84	

 \\
&Qwen7B-Instruct	&  80 	 &		85  &	85 &	--	 &	--	 &	96 		 &	95 		 &42 &	83

 \\
&Qwen72B-Instruct	&85		 &87 	 &	83	 & 	--	 &	--	 &	96 	 &	97 &65	  &83	

\\
&Internlm2.5-7B-Chat&		60 	 &	74  &57	 &	--	 &	--		 &63 &		83  &		27  &		69 

  \\
&Internlm2.5-20B-Chat	&70	 &81 	 &	73 	 &		--		 &--	 &	74  &	83	 &	39 	  & 73	

  \\
 
&LLaMa3.3-70B& 68	 &	81 		 &	54 		 &--	 &	--		 &85  &		97	 &22	 &82

\\

\multirow{-7}{*}{\textbf{Wikipedia}}&	 LongWriter-8B	&73	 &84 &	58  &			--	 &	--		 &97 &		98 	  &	32	 &85	
\\

\midrule[1pt]
\end{tabular}
}
\end{center}
\caption{
The results that GPT4o evaluates on our LongEval benchmark.
}
\label{tab:GPT4o Result}
\end{table*}

In order to demonstrate the reasonability of results evaluated by using Qwen2.5-72B, as shown in Tab~\ref{tab:GPT4o Result}, we also use the GPT4o as the judge model to score the result of different LLMs.

\subsection{Case Study}

In order to better demonstrate the differences in long-text generation among different models under our designed PLAD-based framework, we compare the results generated by InternLM2.5-20B and Qwen2.5-72B, which perform the best and worst, respectively, in the arXiv domain.

\begin{figure*}[!tb]
    \centering
    \includegraphics[width=0.99\textwidth]{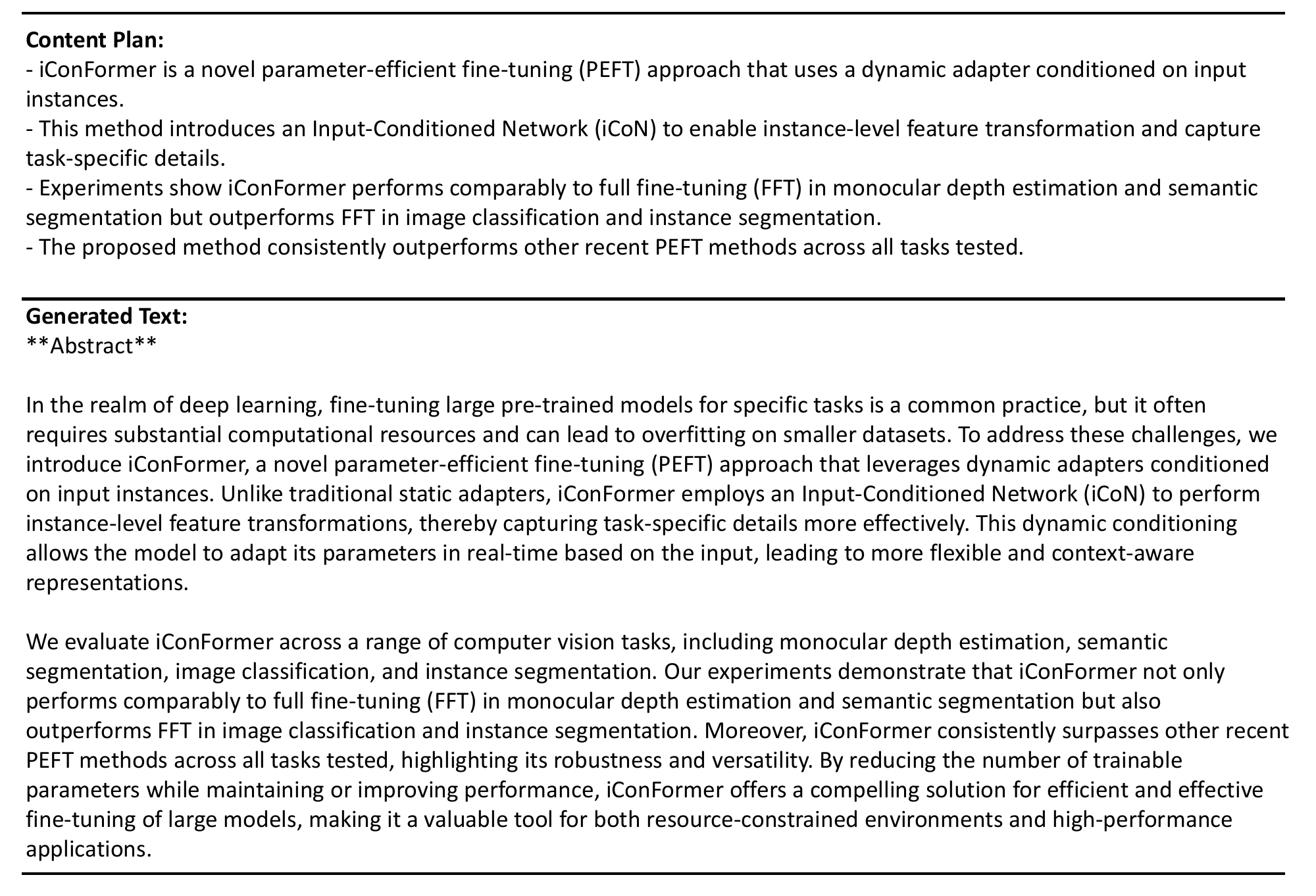}
    \caption{A section generated by Qwen2.5-72B.}
    \label{fig:case1}
\end{figure*}

\begin{figure*}[!tb]
    \centering
    \includegraphics[width=0.99\textwidth]{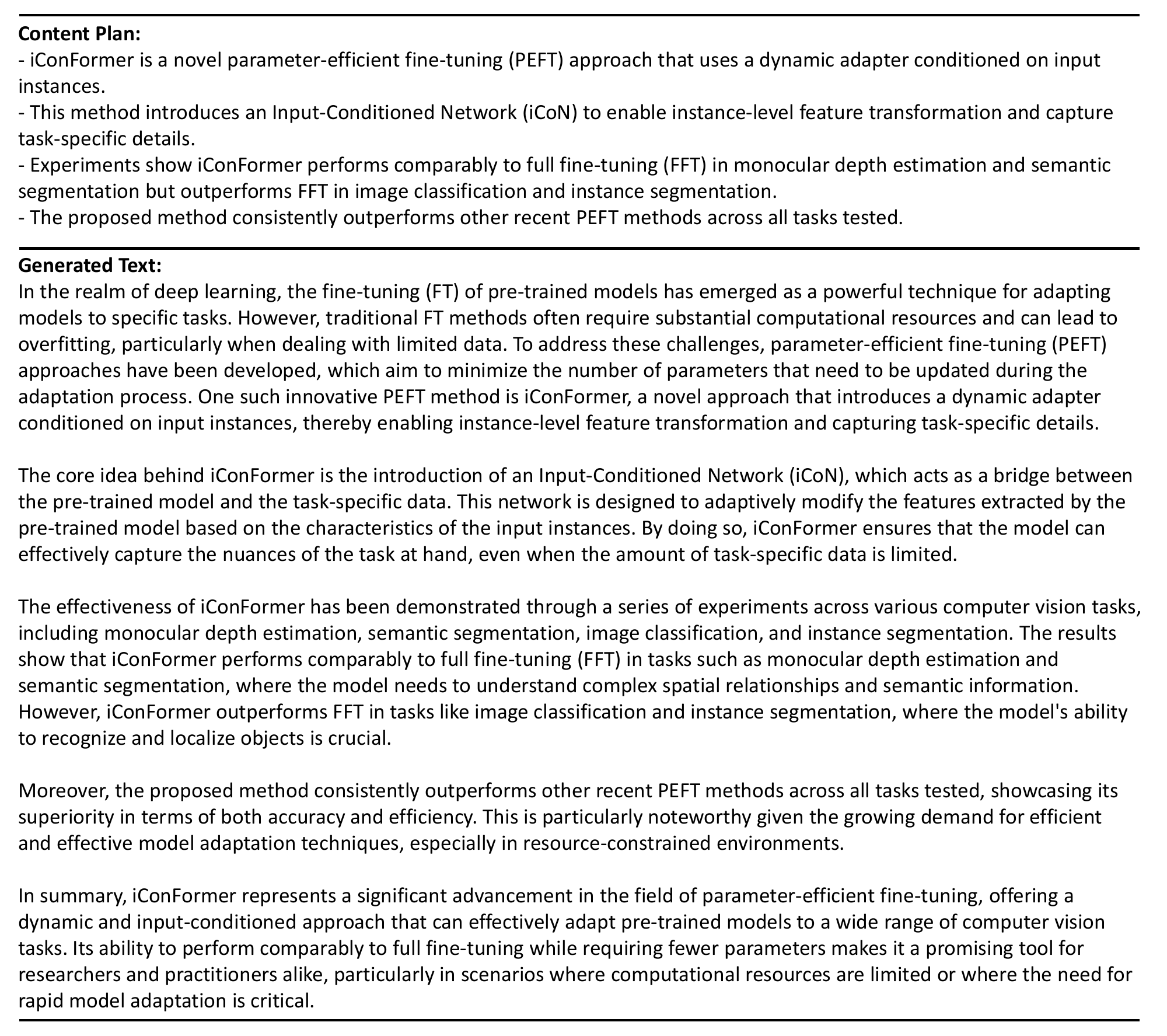}
    \caption{A section generated by InternLM2.5-20B.}
    \label{fig:case2}
\end{figure*}

As shown in Fig~\ref{fig:case1} and Fig~\ref{fig:case2}, for the content plan we summarized, the abstract generated by Qwen2.5 concisely includes all key information and smoothly integrates all content, demonstrating stronger content-following ability. In contrast, the content generated by InternLM2.5-20B is relatively scattered and even includes some unnecessary conclusions at the end, which does not conform to writing conventions for a highly summary-oriented abstract. 


\end{document}